\documentclass{llncs}

\usepackage[subpreambles=true, mode=buildnew]{standalone}
\usepackage[table]{xcolor}
\usepackage{makecell} 
\usepackage{tikz}
\usepackage{graphicx}
\usepackage{subcaption}
 \usepackage{multirow}
 \usepackage{amsmath}
 \usepackage{hyperref}
\usepackage{etoolbox}
\makeatletter
\providecommand{\institute}[1]{
  \apptocmd{\@author}{\end{tabular}
    \par
    \begin{tabular}[t]{c}
    #1}{}{}
}
\makeatother 
\title{Explainable Abstract Trains Dataset}
\author{Manuel de Sousa Ribeiro \and  Ludwig Krippahl \and Joao Leite}
\institute{
NOVA LINCS\\
School of Science and Technology\\
NOVA University Lisbon\\
Portugal}
\date{}
\begin{document}
\newcommand\x[1]{\makebox[#1 pt]{}}

\maketitle

\begin{abstract}
The Explainable Abstract Trains Dataset is an image dataset containing simplified representations of trains. It aims to provide a platform for the application and research of algorithms for justification and explanation extraction.
The dataset is accompanied by an ontology that conceptualizes and classifies the depicted trains based on their visual characteristics, allowing for a precise understanding of how each train was labeled. Each image in the dataset is annotated with multiple attributes describing the trains' features and with bounding boxes for the train elements.
\end{abstract}

\section{Introduction}

\begin{figure}[h!]
	\centering
	\includegraphics{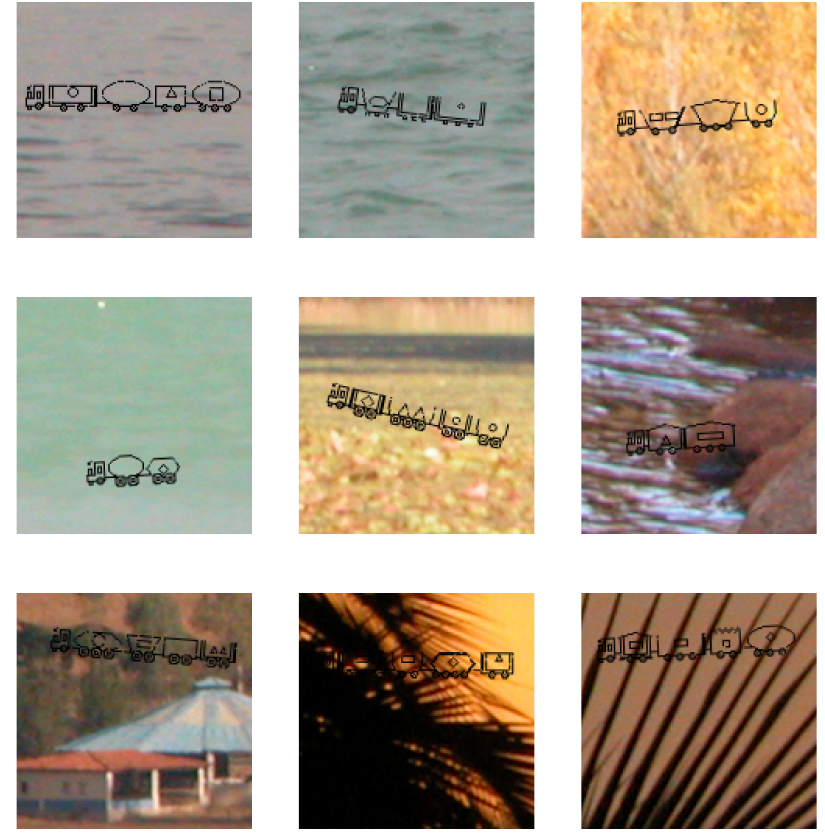}
	\caption{Sample images of trains' representations.}
	\label{fig:trains_and_labels}
\end{figure}

We introduce the Explainable Abstract Trains Dataset\footnote{Available at \url{https://bitbucket.org/xtrains/dataset/}} (XTRAINS), an annotated image dataset focused on explainability, built with the objective of facilitating the research of algorithms for justification and explanation extraction. The XTRAINS dataset contains $500\ 000$ images of $152 \times 152$ pixel art representations of trains, inspired by the trains developed by J. Larson and R. S. Michalski in \cite{Larson1977}, as shown in Figure \ref{fig:trains_and_labels}. 

The XTRAINS dataset was built for and used in \cite{DBLP:conf/aaai/0001RL21} to illustrate a method to produce symbolic justifications for the output of artificial neural networks. It is accompanied by an ontology, shown in Figure \ref{fig:complete_ontology} (in Appendix \ref{onto}), designed to conceptualize and describe the train representations in this dataset. The ontology provides ground-truth knowledge regarding how each image was labeled, allowing for ontology learning methods, such as the ones described in \cite{Ozaki2020} to be benchmarked in this dataset.
The provided bounding boxes allow for attribution methods \cite{Shrikumar2016,Zeiler2014} and salience mapping \cite{Simonyan2013,Zhou2014} methods to be tested, using evaluation metrics, e.g., like the one described in \cite{Schulz2020}.

\section{Image Composition}

Each image in the dataset portrays a single train representation over a background image. All background images are random fragments from the images in the McGill Calibrated Colour Image Database \cite{Olmos2004}. A train is composed by a locomotive and a set of wagons that may have some contents.
Each train representation is characterized as follows:
\begin{itemize}
\item \begin{description} \item[Wagons] The set of wagons contained in the train, further described below. The number of wagons in a train is equal to the integer part of a value sampled from a truncated normal distribution constrained to the interval $[1, 5]$, with a mean of $3$, and a standard deviation of $1$. \end{description} 

\item \begin{description} \item[Wheel size] The size of each wheel in the train, one of $4$ possible sizes. The probability mass function of a train's wheel size is shown in Figure \ref{fig:wheel_size} (in Appendix \ref{app:prob}). \end{description}

\item \begin{description} \item[Couplers' height] The height at which each coupler is attached, one of $2$ possible heights. The height is uniformly distributed between those values. \end{description} 

\item \begin{description} \item[Wagons' spacing] The number of pixels separating each wagon, one of $2$ possible values. The separation is uniformly distributed between those values. \end{description}

\item \begin{description} \item[Position and angle] The trains' position and angle inside an image. A train's position is randomly selected taking into consideration the train's dimensions to ensure that the train is always visible in the image. The angle of each train is sampled from a truncated normal distribution constrained to the interval $[-30, 30]$, with a mean of $0$, and a standard deviation of $9$.
\end{description}
\end{itemize}

\begin{figure}[h]
	\centering
	\includegraphics[scale=0.77]{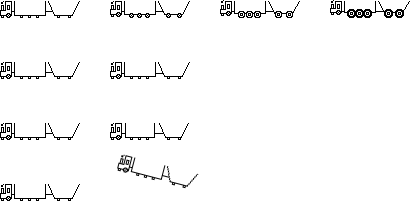}
	\caption{Effect of individually varying trains' features. 
	}
	\label{fig:trains_features}
\end{figure}

The effect of varying the wheel size, couplers' height, wagons' spacing and the position and angle of a train is shown, respectively, in each row of Figure \ref{fig:trains_features}.

A wagon is characterized as follows:
\begin{itemize}
\item \begin{description} \item[Length] The wagon's length in pixels. A wagon has a length of $34$ pixels with a probability of $1/3$ and a length of $24$ pixels with a probability of $2/3$.  \end{description}

\item \begin{description} \item[Wall shape] The shape of the wagon's walls, one of $6$ available shapes. \end{description}

\item \begin{description} \item[Roof shape] The shape of the wagon's roof, one of $5$ available shapes. \end{description}

\item \begin{description} \item[Amount of visible wheels] The number of visible wheels in a wagon, one of $3$ possible values. \end{description}

\item \begin{description} \item[Contents] The contents carried by a wagon, further described below. \end{description}
\end{itemize}

\begin{figure}[h]
	\centering
	\includegraphics{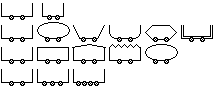}
	\caption{Effect of individually varying wagons' features. First row varies length; second row varies wall shape; third row varies roof shape; fourth row varies the amount of visible wheels.}
	\label{fig:wagons_features}
\end{figure}

The values of a wagons' wall shape, roof shape and amount of wheels are conditioned by the wagon's length and by each others' values, with probability mass functions of these features presented, respectively, in Figure \ref{fig:wall_shape}, \ref{fig:roof_shape} and \ref{fig:number_of_wheels} (in Appendix \ref{app:prob}). The effect of individually varying each feature is depicted in Figure \ref{fig:wagons_features}.

The contents of a wagon, if any, are characterized as follows:
\begin{itemize}
\item \begin{description} \item[Shape] The shape of the content, one of $8$ available shapes. \end{description}

\item \begin{description} \item[Size] The size of the content, one of $4$ available sizes. \end{description}

\item \begin{description} \item[Quantity] The content's quantity, one of $3$ possible values. \end{description}

\item \begin{description} \item[Position] The position of the contents inside a wagon. This value is dependent on the contents' shape, size, and quantity and on the trains' length and wheel size, ensuring that the contents are always inside the wagon.  \end{description}
\end{itemize}

\begin{figure}[h]
	\centering
	\includegraphics{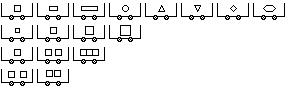}
	\caption{Effect of individually varying wagons' contents' features. First row varies shape; second row varies size; third row varies quantity; fourth row varies position.}
	\label{fig:content_features}
\end{figure}

The probabilities associated with the values of the shape, size, and quantity of a wagons' content are conditioned by the wagon's length, wall shape, and roof shape and by each others' values, and can be seen, respectively, in Figures \ref{fig:content_shape}, \ref{fig:content_size}, and \ref{fig:content_quantity} (in Appendix \ref{app:prob}). The effect of varying each of these features individually is illustrated, respectively, in each row of Figure \ref{fig:content_features}. 

\begin{figure}[h]
	\centering
	\includegraphics{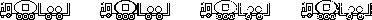}
	\caption{Addition of noise to the leftmost train representation, the second and third trains show, respectively, the effect of individually adding the possibility of thick walls and missing pixels to the first train.
	}
	\label{fig:noises}
\end{figure}

Additionally, we associate to the floor and each wall of a wagon a $15\%$ probability of it being thicker, drawing it with a thickness of $2$ pixels, instead of $1$. Noise was also deliberately introduced in the form of missing pixels from the trains' representations, up to $10\%$ of the trains' pixels might not be drawn. The resulting effect is depicted in Figure \ref{fig:noises}, the first train has no thick walls or missing pixels and the fourth train shows both effects combined. 

\section{Dataset Annotations}

All images in the dataset are annotated with $39$ binary attributes, shown in Figure \ref{fig:binary_attributes} (in Appendix \ref{app:attr}), such as $\sf TwoWheelsWagon$ indicating if there is any wagon with $2$ wheels visible in the image, and with $8$ numeric attributes, shown in Figure \ref{fig:numeric_attributes} (in Appendix \ref{app:attr}), like $\sf Angle$ indicating the angle of the train inside an image. The values for those attributes were either obtained directly when generating each image, e.g., the value of the attribute $\sf NumberOfWagons$, indicating the number of wagons in an image, or obtained through reasoning with the axioms in the datasets' ontology, e.g., $\sf LongTrain$, indicating a train with, at least, $2$ long wagons or $3$ wagons. Furthermore, each image was also annotated with bounding boxes for the following elements:
\begin{itemize}
\item Train;
\item Train's Wagon;
\item Wagon's Contents;
\item Wagon's Roof;
\item Wagon's Wheels.
\end{itemize}

\section{Image Generator}
We also make available an image generator\footnote{Available at \url{https://bitbucket.org/xtrains/dataset/}} capable of generating the XTRAINS dataset. This generator has multiple parameters, e.g., to define the length of the trains' wagons, or to define the different possible amounts of wheels of a wagon, which are set to be in accord with the descriptions provided above. By changing the values of these parameters, it is possible to generate new datasets with different characteristics. 

\section{Conclusion}

The XTRAINS dataset has a total $500\ 000$ images over multiple different classes of trains. We hope that the large number of different annotations provided, coupled together with the knowledge embedded in the provided ontology make the XTRAINS dataset useful to benchmark different methods and techniques, notably in the area of Explainable AI.

\subsubsection*{Acknowledgements}
The authors would like to thank the support provided by Calouste Gulbenkian Foundation through its \emph{Young Talents in AI} program, by FCT project ABSOLV (PTDC/CCI-COM/28986/2017), and by FCT strategic project NOVA LINCS (UIDB/04516/2020).

\bibliographystyle{plain}
\bibliography{xtrains}

\begin{thebibliography}{1}

\bibitem{DBLP:conf/aaai/0001RL21}
Manuel de~Sousa~Ribeiro and Joao Leite.
\newblock Aligning artificial neural networks and ontologies towards
  explainable {AI}.
\newblock In {\em The Thirty-Fifth {AAAI} Conference on Artificial
  Intelligence, {AAAI'21}}. {AAAI} Press, 2021.

\bibitem{Larson1977}
J.~Larson and Ryszard~S. Michalski.
\newblock Inductive inference of {VL} decision rules.
\newblock {\em {SIGART} Newsl.}, 63:38--44, 1977.

\bibitem{Olmos2004}
Adriana Olmos and Frederick A.~A. Kingdom.
\newblock A biologically inspired algorithm for the recovery of shading and
  reflectance images.
\newblock {\em Perception}, 33(12):1463--1473, 2004.
\newblock PMID: 15729913.

\bibitem{Ozaki2020}
Ana Ozaki.
\newblock Learning description logic ontologies: Five approaches. where do they
  stand?
\newblock {\em K{\"{u}}nstliche Intell.}, 34(3):317--327, 2020.

\bibitem{Schulz2020}
Karl Schulz, Leon Sixt, Federico Tombari, and Tim Landgraf.
\newblock Restricting the flow: Information bottlenecks for attribution.
\newblock In {\em 8th International Conference on Learning Representations,
  {ICLR} 2020, Addis Ababa, Ethiopia, April 26-30, 2020}. OpenReview.net, 2020.

\bibitem{Shrikumar2016}
Avanti Shrikumar, Peyton Greenside, Anna Shcherbina, and Anshul Kundaje.
\newblock Not just a black box: Learning important features through propagating
  activation differences.
\newblock {\em CoRR}, abs/1605.01713, 2016.

\bibitem{Simonyan2013}
Karen Simonyan, Andrea Vedaldi, and Andrew Zisserman.
\newblock Deep inside convolutional networks: Visualising image classification
  models and saliency maps.
\newblock In Yoshua Bengio and Yann LeCun, editors, {\em 2nd International
  Conference on Learning Representations, {ICLR} 2014, Banff, AB, Canada, April
  14-16, 2014, Workshop Track Proceedings}, 2014.

\bibitem{Zeiler2014}
Matthew~D. Zeiler and Rob Fergus.
\newblock Visualizing and understanding convolutional networks.
\newblock In David~J. Fleet, Tom{\'{a}}s Pajdla, Bernt Schiele, and Tinne
  Tuytelaars, editors, {\em Computer Vision - {ECCV} 2014 - 13th European
  Conference, Zurich, Switzerland, September 6-12, 2014, Proceedings, Part
  {I}}, volume 8689 of {\em Lecture Notes in Computer Science}, pages 818--833.
  Springer, 2014.

\bibitem{Zhou2014}
Bolei Zhou, Aditya Khosla, {\`{A}}gata Lapedriza, Aude Oliva, and Antonio
  Torralba.
\newblock Object detectors emerge in deep scene cnns.
\newblock In Yoshua Bengio and Yann LeCun, editors, {\em 3rd International
  Conference on Learning Representations, {ICLR} 2015, San Diego, CA, USA, May
  7-9, 2015, Conference Track Proceedings}, 2015.

\end{thebibliography}

\newpage

\appendix
\section{Ontology}
\label{onto}

\begin{figure*}[!h]
	\centering
	\makebox[0pt]{
	\includegraphics{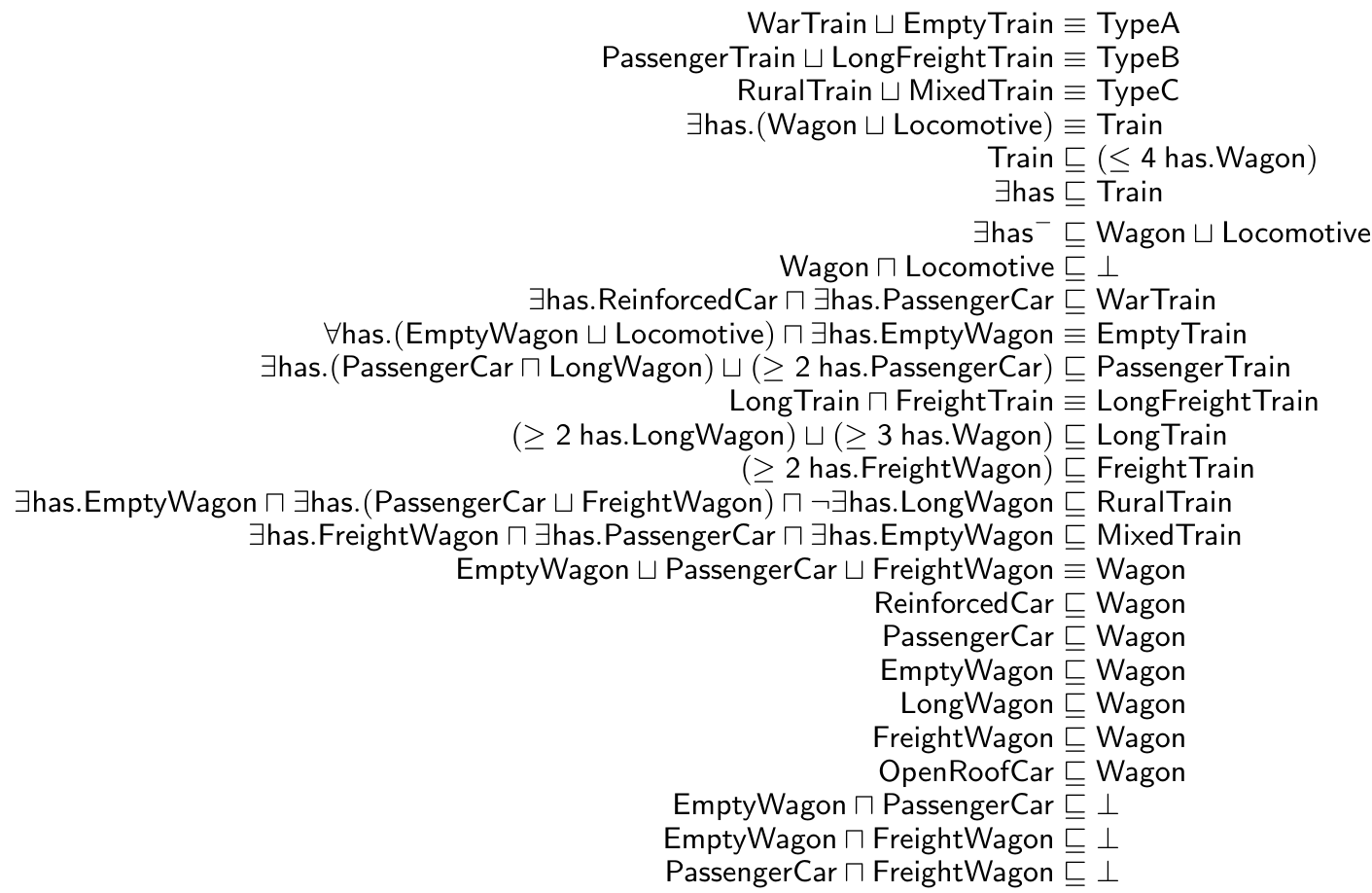}
	}
    \caption{Ontology describing how the trains' representations are classified.}
    \label{fig:complete_ontology}
\end{figure*}

\clearpage
\newpage

\section{Train Features' Probability Mass Function}
\label{app:prob}

\begin{figure*}[!h]
	\centering
	\includegraphics{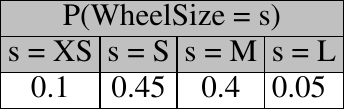}
    \caption{Probability mass function of a train's wheel size.}
    \label{fig:wheel_size}
\end{figure*}

\begin{figure*}[!h]
	\centering
	\includegraphics{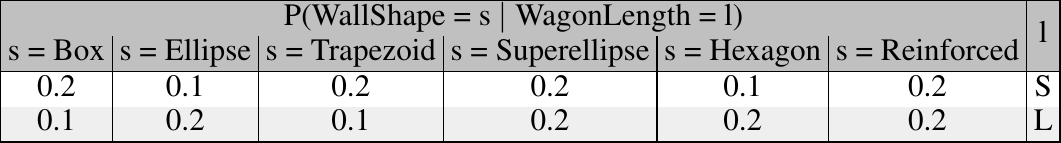}
    \caption{Probability mass function of a wagon's wall shape.}
    \label{fig:wall_shape}
\end{figure*}

\begin{figure*}[!h]
	\centering
	\includegraphics{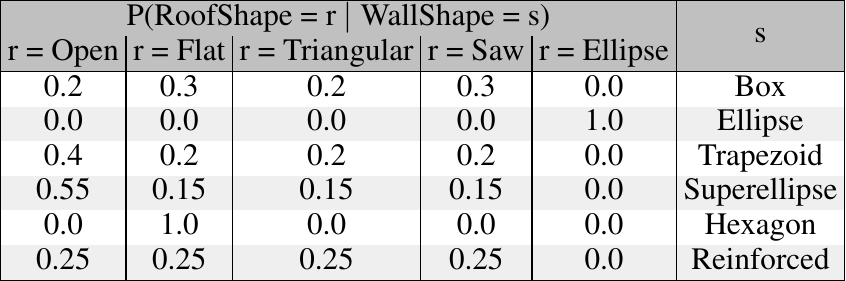}
    \caption{Probability mass function of a wagon's roof shape.}
    \label{fig:roof_shape}
\end{figure*}

\begin{figure*}[!h]
	\centering
	\includegraphics{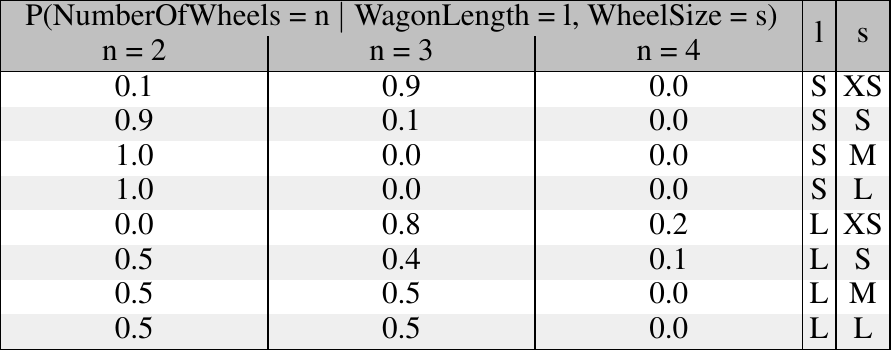}
    \caption{Probability mass function of a wagon's number of visible wheels.}
    \label{fig:number_of_wheels}
\end{figure*}

\begin{figure*}[!h]
	\centering
	\makebox[0pt]{
	\includegraphics{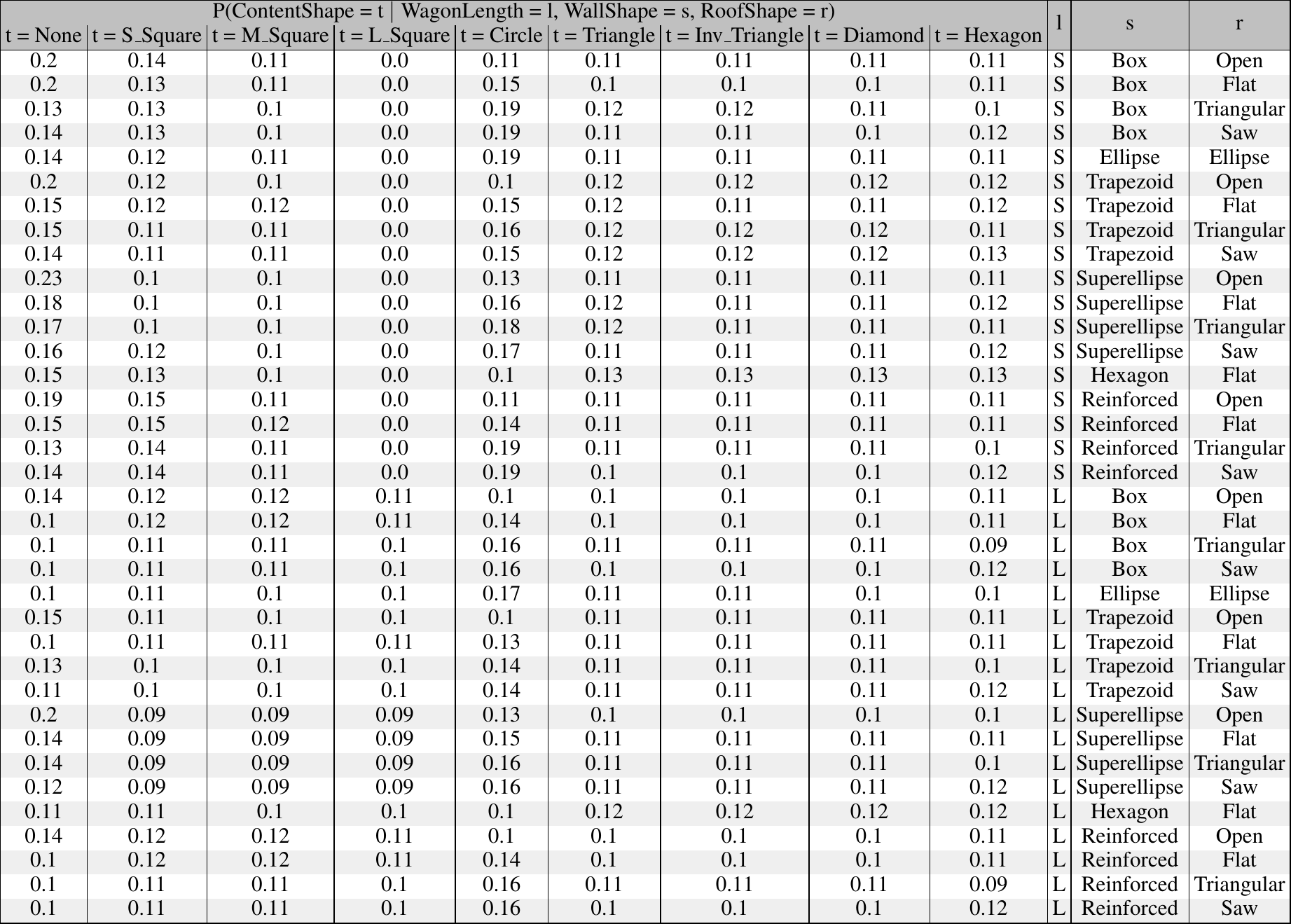}
	}
    \caption{Probability mass function of a wagon's content shape.}
    \label{fig:content_shape}
\end{figure*}

\begin{figure*}[!h]
	\centering
	\makebox[0pt]{
	\includegraphics{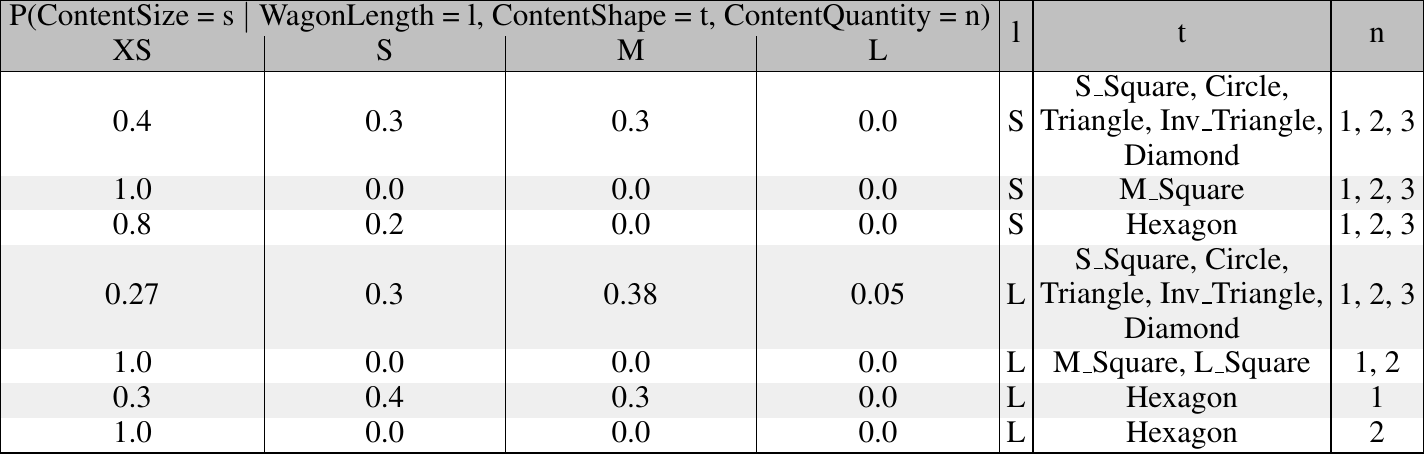}
	}
    \caption{Probability mass function of a wagon's content size.}
    \label{fig:content_size}
\end{figure*}

\begin{figure*}[!h]
	\centering
	\makebox[0pt]{
	\includegraphics{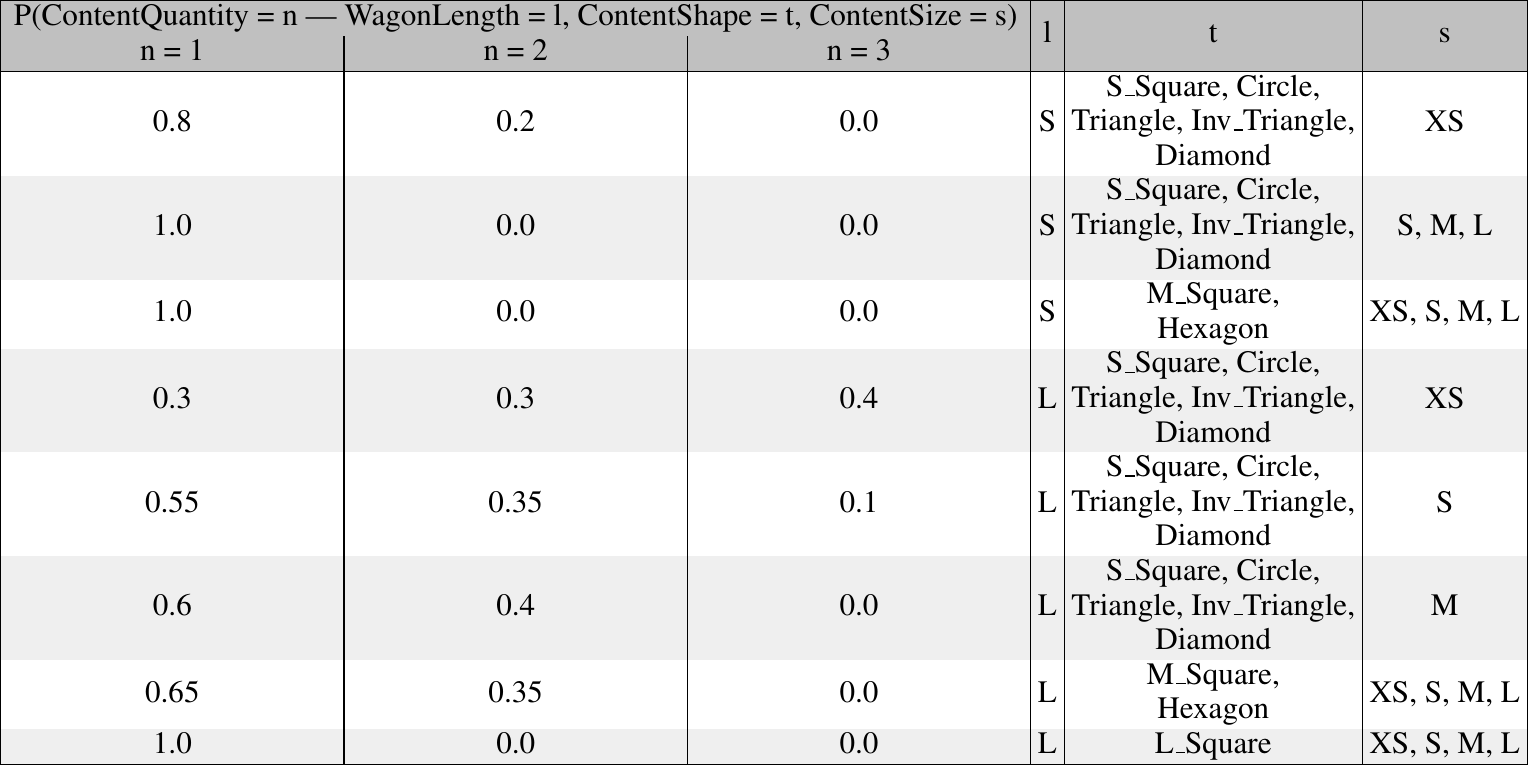}
	}
    \caption{Probability mass function of a wagon's content quantity.}
    \label{fig:content_quantity}
\end{figure*}

\clearpage
\newpage

\section{Dataset Attributes}
\label{app:attr}

\begin{figure*}[!h]
	\centering
	\makebox[0pt]{
	\includegraphics{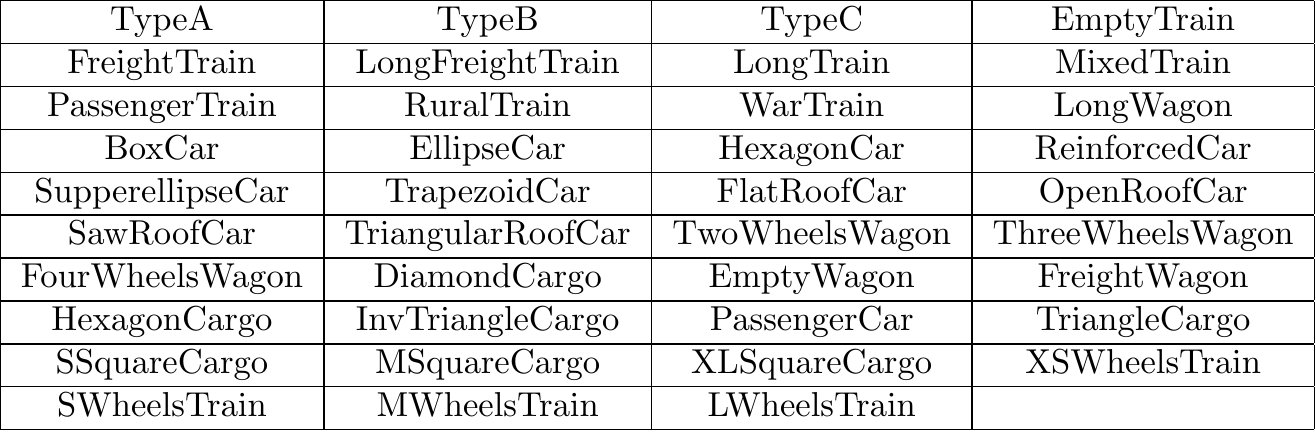}
	}
    \caption{XTRAINS dataset's binary attributes.}
    \label{fig:binary_attributes}
\end{figure*}

\begin{figure*}[!h]
	\centering
	\makebox[0pt]{
	\includegraphics{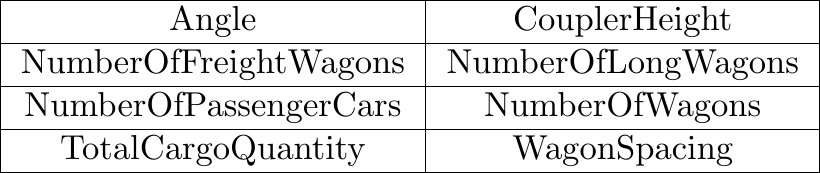}
	}
    \caption{XTRAINS dataset's numeric attributes.}
    \label{fig:numeric_attributes}
\end{figure*}

\end{document}